\documentclass[lettersize,journal]{IEEEtran}
\usepackage{amsmath,amsfonts}
\usepackage{algorithmic}
\usepackage{algorithm}
\usepackage{array}
\usepackage[caption=false,font=normalsize,labelfont=sf,textfont=sf]{subfig}
\usepackage{textcomp}
\usepackage{stfloats}
\usepackage{url}
\usepackage{verbatim}
\usepackage{graphicx}
\usepackage{cite}
\usepackage{float}
\usepackage{booktabs}
\usepackage{multirow}
\usepackage{fancyhdr}
\usepackage{marvosym}
\newcommand{\ie}{\textit{i.e.}}

\begin{document}

\title{Identity-Preserving Video Dubbing Using Motion Warping}

\author{Runzhen Liu,  Qinjie Lin, Yunfei Liu, Lijian Lin, Ye Zhu, Yu Li, Chuhua Xian{\textsuperscript{\Letter}}, Fa-Ting Hong{\textsuperscript{\Letter}}

\thanks{Runzhen Liu and Chuhua Xian are with the Department of Computer Science and Engineering, South China University of Technology. E-mail: csalunaticat@mail.scut.edu.cn, chhxian@scut.edu.cn.}

\thanks{Qinjie Lin is with Department of Computer Science, Northwestern University, USA. E-mail:
qinjielin2018@u.northwestern.edu.}

\thanks{Yunfei Liu, Lijian Lin, Ye Zhu, Yu Li are with Vistring Lab, IDEA. E-mail: liuyunfei@idea.edu.cn, ljlin@stu.xmu.edu.cn, zhu\_ye\_ye@outlook.com, liyu@idea.edu.cn.}

\thanks{Fa-Ting Hong is with the Department of Computer Science and Engineering, The Hong Kong University of Science and Technology, Hong Kong SAR. E-mail: fhongac@connect.ust.hk.}
}

\markboth{Journal of \LaTeX\ Class Files,~Vol.~14, No.~8, August~2021}%
{Shell \MakeLowercase{\textit{et al.}}: A Sample Article Using IEEEtran.cls for IEEE Journals}


\maketitle

\begin{abstract}
Video dubbing aims to synthesize realistic, lip-synced videos from a reference video and a driving audio signal. Although existing methods can accurately generate mouth shapes driven by audio, they often fail to preserve identity-specific features, largely because they do not effectively capture the nuanced interplay between audio cues and the reference identity’s visual attributes. As a result, the generated outputs frequently lack fidelity in reproducing the unique textural and structural details of the reference identity. 
To address these limitations, we propose \textbf{IPTalker}, a novel and robust framework for video dubbing that achieves seamless alignment between driving audio and reference identity while ensuring both lip-sync accuracy and high-fidelity identity preservation. At the core of IPTalker is a transformer-based alignment mechanism designed to dynamically capture and model the correspondence between audio features and reference images, thereby enabling precise, identity-aware audio-visual integration.
Building on this alignment, a motion warping strategy further refines the results by spatially deforming reference images to match the target audio-driven configuration. A dedicated refinement process then mitigates occlusion artifacts and enhances the preservation of fine-grained textures, such as mouth details and skin features. Extensive qualitative and quantitative evaluations demonstrate that IPTalker consistently outperforms existing approaches in terms of realism, lip synchronization, and identity retention, establishing a new state of the art for high-quality, identity-consistent video dubbing. 
\end{abstract}


\section{Introduction}

\IEEEPARstart{V}{ideo} dubbing, a technique focused on generating realistic facial videos from a reference video and a driving audio signal, has emerged as a pivotal tool in domains such as augmented and virtual reality (AR/VR), video conferencing, digital animation, film production, and other entertainment sectors. This technology holds the potential to reshape how humans engage with and experience multimedia content.

At the core of video dubbing lies the challenge of multimodal fusion, requiring precise integration of audio and visual information. The audio component dictates dynamic elements such as mouth shapes and movements, while visual features capture static attributes like facial expressions, skin tone, eye color, and mouth texture details. By learning the intricate mapping between audio signals and corresponding mouth movements, a video dubbing model can generate lifelike videos that preserve the reference identity while faithfully adhering to the driving audio.

Numerous methods have been proposed to tackle video dubbing, with most focusing on audio-to-mouth synchronization. Direct pixel-based synthesis approaches, such as Wav2Lip~\cite{prajwal2020lip}, employ pre-trained lip-sync discriminators to guide the generation of facial frames, whereas landmark-based techniques~\cite{zhong2023identity, zhou2020makelttalk, xie2021towards} predict landmarks from the audio input and then render images based on those predictions. Although these methods excel at creating accurate lip movements, they often overlook the interplay between the reference identity and driving audio during feature extraction, resulting in outputs that do not faithfully preserve the unique facial characteristics of the reference identity.

\IEEEpubidadjcol

\begin{figure}[t]
  \centering
  \includegraphics[width=\linewidth]{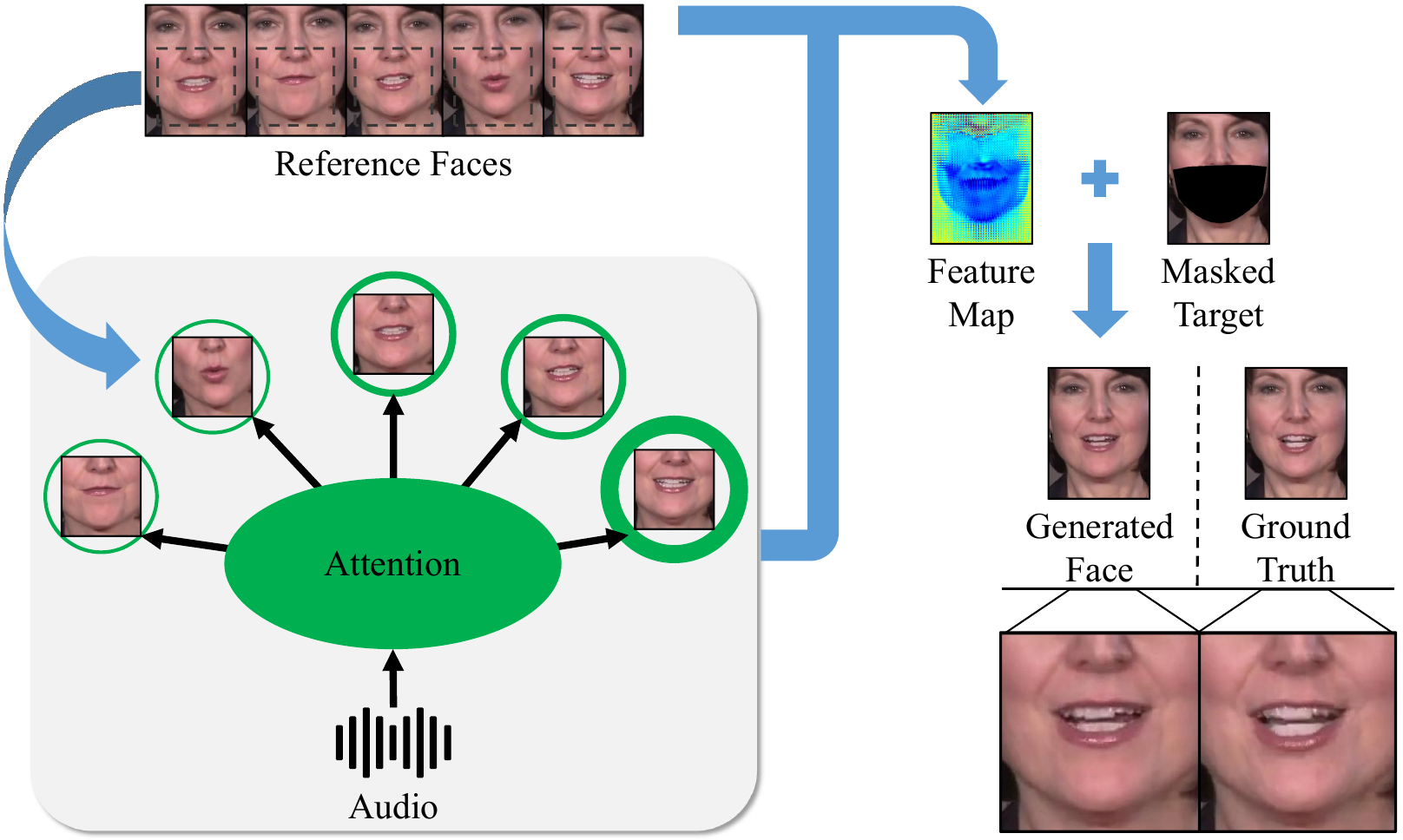}
  \vspace{-17pt}
  \caption{Our proposed IPTalker generates dubbed videos using a few reference images to provide identity-specific priors and driving audio to dictate mouth shapes. The synthesis is achieved by inpainting the masked source image, ensuring rich textural details and temporal consistency.}
  \vspace{-15pt}
  \label{fig:introduction}
\end{figure}

In the realm of video dubbing, direct pixel-based methods~\cite{kr2019towards, prajwal2020lip, xie2021towards} often simplify the synthesis process but struggle to produce high-resolution outputs. Additionally, their limited capacity to model fine-grained correlations between driving audio and the mouth’s texture constrains their effectiveness. Neural Radiance Field (NeRF)-based methods~\cite{liu2022semantic, shen2022learning, guo2021ad} leverage neural rendering to generate facial images by learning scene-specific representations; however, they typically require a separate NeRF model for each identity, making them computationally expensive and less scalable. GAN-based techniques~\cite{prajwal2020lip, liang2022expressive} can generate identity-consistent facial images but often suffer from temporal inconsistencies when relying solely on audio cues, compromising the realism of the final output.

To address these challenges and improve both identity preservation and textural fidelity, we propose \textbf{IPTalker}, a novel framework for audio-driven video dubbing. The core innovation of IPTalker lies in its effective use of reference images, which are aligned and deformed to synthesize temporally consistent and visually compelling results.

Since the audio signal and mouth shape share a strong correspondence (e.g., the mouth opens when pronouncing ``A''), we introduce the Audio-Visual Alignment Unit (AVAU), a transformer-based component that learns a correspondence embedding between audio features and reference images. This embedding pinpoints the reference image most closely aligned with the target mouth shape dictated by the driving audio. By explicitly modeling this audio-visual interplay and incorporating a spatial deformation mechanism, IPTalker better preserves the fine-grained textural details of the reference identity. Furthermore, a Spatial Deformation Module warps reference images according to the predicted motion flow to maintain fidelity to both the identity and driving audio. Finally, an Inpainting Module resolves occlusions or artifacts resulting from the warping step, refining the overall visual quality of generated frames.

We conduct extensive experiments on two public datasets, \ie, VFHQ~\cite{xie2022vfhq} and HDTF~\cite{zhang2021flow}, to evaluate the performance of IPTalker. Experimental results indicate that our method achieves state-of-the-art performance in generating realistic videos while preserving the identity information of the reference subject. The contributions of this work are summarized as follows:
\begin{itemize}
    \item We investigate the nuanced relationship between audio signals and mouth shapes, highlighting how variations in phonemes govern the dynamic movements of the lips and adjacent facial regions. By modeling this correspondence, IPTalker effectively integrates the driving audio with reference visual features, capturing subtle temporal and spatial cues.
    \item We propose IPTalker, a holistic framework for audio-driven video dubbing that produces realistic facial animations with high textural fidelity. An alignment module identifies the reference mouth images best suited for the driving audio, and a warping module exploits the learned audio-visual cues to generate motion flows that warp the reference images.
    \item Through both qualitative and quantitative evaluations on the HDTF~\cite{zhang2021flow} and VFHQ~\cite{xie2022vfhq} datasets, we demonstrate that IPTalker outperforms existing approaches in realism, temporal consistency, identity preservation, and lip-sync accuracy.
\end{itemize}

By tackling critical challenges in identity preservation and multimodal fusion, IPTalker establishes a new benchmark in video dubbing and offers a robust solution for diverse applications in AR/VR, entertainment, and communication technologies.

\section{RELATED WORK}

\subsection{Audio-Driven Talking Head Generation}
Existing audio-driven talking head generation methods can be categorized into 3D-based and 2D-based approaches. 

\textbf{3D-Based Methods}: A common intermediate representation in 3D-based methods is the 3D Morphable Model (3DMM)~\cite{2002A}, which decomposes facial attributes such as shape, expression, and appearance, facilitating the modification of specific attributes. For instance, \textit{StyleTalk}~\cite{ma2023styletalk} utilizes 3DMM to extract expression parameters for synthesizing stylized faces. However, existing 3DMM models often lack accuracy and struggle to generate detailed facial textures. Another prominent 3D approach is Neural Radiance Fields (NeRF)~\cite{mildenhall2021nerf}, which learns to map images from various angles to a 3D model using neural networks~\cite{liu2022semantic, shen2022learning, guo2021ad}. For example, \textit{AD-NeRF}~\cite{guo2021ad} constructs separate NeRF models for the head and torso, enabling pose controllability. Despite their capabilities, NeRF-based methods are computationally intensive and typically require training a separate model for each identity, limiting their scalability for large-scale applications.

\textbf{2D-Based Methods}: 2D-based approaches often generate images directly from reference images and driving audio. \textit{Wav2Lip}~\cite{prajwal2020lip} generates talking heads by utilizing audio Mel-spectrograms and source images, supervised by a pre-trained lip-sync discriminator. Many 2D-based methods employ facial landmarks as intermediate representations. For example, \textit{MakeItTalk}~\cite{zhou2020makelttalk} predicts a sequence of landmarks based on speech content and speaker identity, which are then used to render the corresponding facial images. Similarly, \textit{IPLAP}~\cite{zhong2023identity} predicts lip and jaw landmarks using audio, reference landmarks, and pose priors through self-attention mechanisms. The predicted landmarks are combined with pose priors to generate target sketches, which are subsequently used to produce the talking head.

Despite the advancements, these methods often fail to effectively leverage the relationship between the reference image's lips and the driving audio. We argue that by minimizing the feature distance between the driving audio and the target lip shape, the reference mouth image with the smallest feature distance is most closely aligned with the desired mouth configuration. Assigning different weights to reference mouth images based on their feature distances allows the model to focus on high-weight references during the warping process, facilitating the generation of accurate target mouth images. This weighted approach also ensures that the texture details of high-weight reference images are prominently featured, while those of low-weight references are appropriately filtered, thereby preserving the integrity of the generated mouth textures.

\begin{figure*}[ht]
    \centering
    \includegraphics[width=\textwidth]{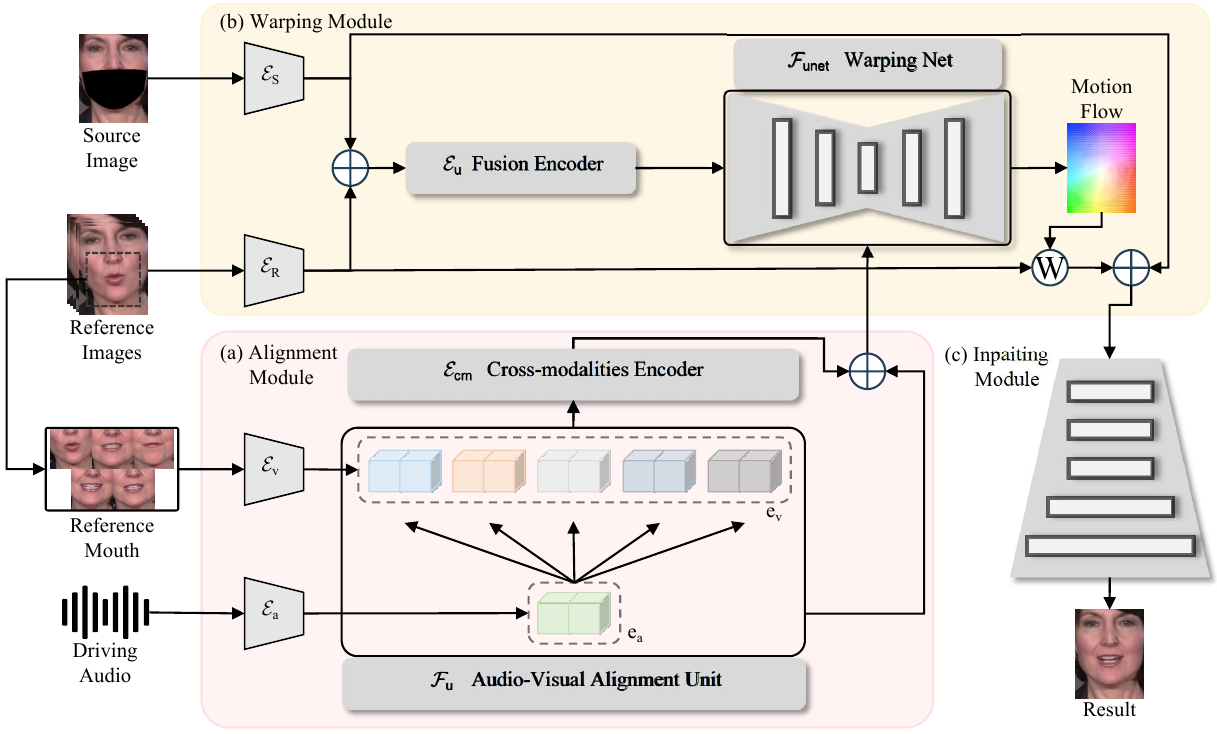}
    \caption{The framework of our method consists of three components: \textbf{(a) Alignment Module}, where reference mouth images and driving audio are input into encoders to extract embeddings. The Audio-Visual Alignment Unit (AVAU) captures the relationships among all embeddings to obtain an identity-audio correspondence embedding. \textbf{(b) Warping Module}, which uses the reference image and the identity-audio correspondence embedding to generate a motion flow that deforms the reference image to match the target configuration dictated by the audio. \textbf{(c) Inpainting Module}, which inpaints the masked source image to produce the final generated image.}
    \label{fig:framework}
\end{figure*}

\subsection{Spatial Deformation}
Some audio-driven talking head methods generate facial images directly from pixels~\cite{kr2019towards, prajwal2020lip, xie2021towards}. For example, \textit{Wav2Lip}~\cite{prajwal2020lip} employs a convolutional decoder to generate mouth images directly from audio Mel-spectrograms and source images using a pre-trained lip-sync discriminator. However, these direct-generation-based methods often struggle to produce high-resolution videos with rich mouth texture details. With only a limited number of reference images, it is challenging for the model to learn a direct mapping from a wide variety of face shapes and speaking styles to corresponding mouth movements. Moreover, while audio primarily influences the shape and movement of the lips, it has little impact on finer details such as lip color, tooth shape and color, or tongue positioning. Minor discrepancies in these details can significantly reduce the realism of the generated videos.

To address these issues, instead of generating images directly from pixels, our method learns the feature distance between the reference and target images using self-attention~\cite{vaswani2017attention} and warps the reference mouth images to align with the target mouth shapes through spatial deformation. This approach preserves the authentic texture details of the reference images. Spatial deformation can be achieved through various techniques, including affine transformations and optical flow.

\textbf{Affine Transformation-Based Methods}: Approaches like \textit{DINet}~\cite{zhang2023dinet} utilize the AdaAT~\cite{zhang2022adaptive} operator to simulate complex deformations by calculating different affine coefficients for different feature channels.

\textbf{Optical Flow-Based Methods}: Methods such as \textit{PIRenderer}~\cite{ren2021pirenderer} employ a warping network with an auto-encoder architecture that takes an image and a latent vector as input. The network uses AdaIN~\cite{huang2017arbitrary} operators to inject the target motion descriptor after each convolutional layer, ultimately producing a flow field to warp the reference image to the target image. \textit{FOMM}~\cite{siarohin2019first} combines sparse optical flow with local affine transformations to generate dense optical flow, using an occlusion mask to distinguish between regions that can be warped and those that require inpainting.

Our method leverages a flow field to warp reference images, ensuring that the mouth texture details of the reference images are well-preserved in the generated target images. By accurately modeling the spatial deformation, IPTalker maintains high fidelity to both the identity and the audio-driven dynamics, resulting in more realistic and temporally consistent video outputs.

\section{Method}

In this work, we propose a novel framework, \textbf{IPTalker}, designed to preserve identity information in video dubbing. We utilize an alignment module to select appropriate mouth appearance information by investigating the correspondence between audio and lip motion. This enables IPTalker to produce accurately lip-synced faces while maintaining identity similarity.

\subsection{Overview}

Given a source video and a driving audio input, our goal is to achieve visually dubbed faces with rich textural details. Consequently, we introduce \textbf{IPTalker} (illustrated in Figure~\ref{fig:framework}), an identity-preserving video dubbing framework comprising three main components:

\begin{enumerate}
    \item \textbf{Alignment Module}: To obtain more appropriate appearance features, the alignment module utilizes audio and a set of lower-half facial images to generate an identity-audio correspondence embedding within a transformer block. This ensures that the subsequent generation process is aware of both the audio information and the texture of the lower half of the face.
    
    \item \textbf{Warping Module}: Given a reference image containing identity information and the identity-audio correspondence embedding, the warping module is designed to produce a motion flow that maps the desired target image to the given reference images.
    
    \item \textbf{Inpainting Module}: After warping the reference image with the generated motion flow, the inpainting module compensates for occluded regions not addressed by the motion flow, effectively inpainting the masked source image.
\end{enumerate}

\begin{figure*}[t]
\centering
\includegraphics[width=\linewidth]{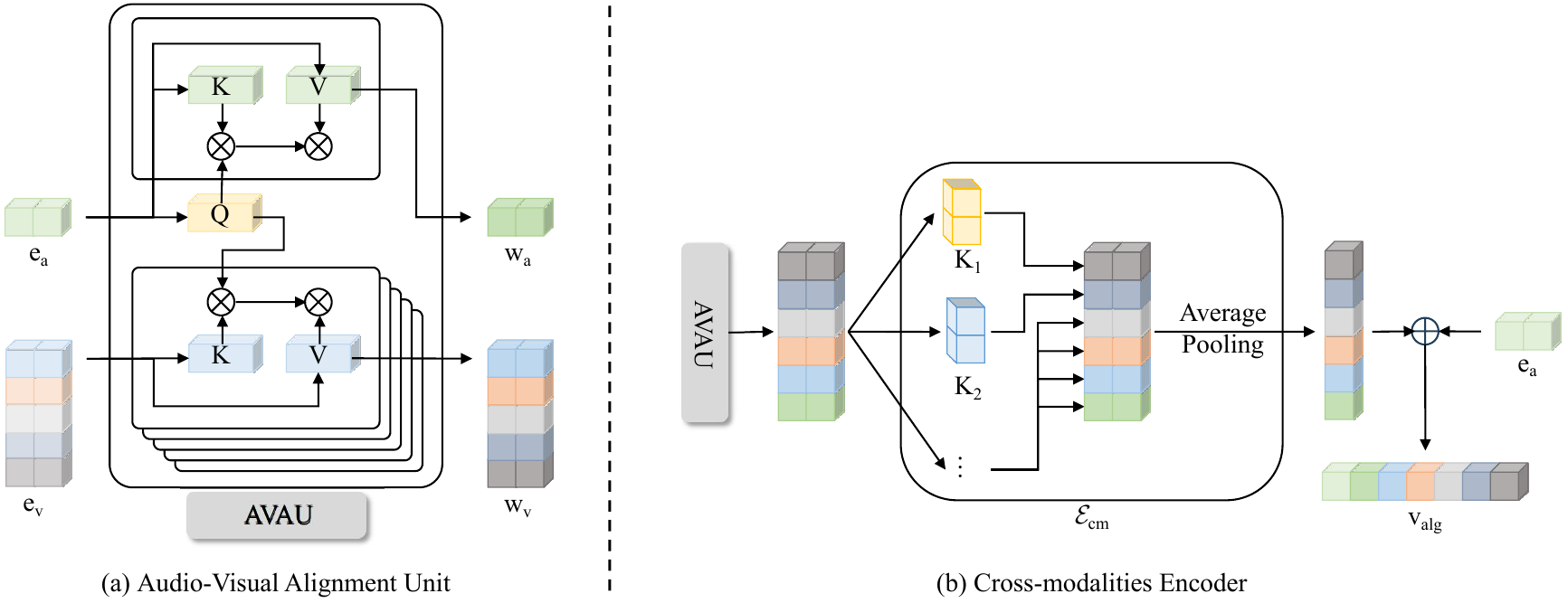}
\caption{Illustration of the Audio-Visual Alignment Unit (AVAU) and the cross-modalities encoder $\mathcal{E}_{cm}$. The AVAU captures the intricate interplay between audio and visual embeddings through self-attention and cross-attention mechanisms (see a). The cross-modalities encoder $\mathcal{E}_{cm}$ compresses the output of multiple AVAUs using 1D convolution and average pooling (see b).}
\label{fig:AVAU_align}
\end{figure*}

\subsection{Alignment Module}

Capturing the intricate relationship between audio signals and lip movements is crucial for talking head generation. While direct mapping approaches ensure high accuracy in replicating mouth shapes corresponding to spoken audio, they often fall short in preserving essential reference information, such as detailed mouth textures including tongue shape and tooth color. These nuances are vital for maintaining the identity of the reference person. To address this limitation and enhance the realism of generated talking heads, incorporating reference images into the alignment module is a promising strategy.

\noindent\textbf{Multi-Modality Extraction.}  
We input two modalities into our alignment module. Initially, we employ a visual encoder $\mathcal{E}_v$ and an audio encoder $\mathcal{E}_a$ to extract visual and audio embeddings, denoted as $\{\mathbf{e}_v^i\}_{i=1}^{N}$ and $\mathbf{e}_a$, from the input images and audio, respectively:
\begin{equation}
\begin{aligned}
    \mathbf{e}_a &= \mathcal{E}_a(\mathbf{A}) + \mathbf{e}_\alpha, \\
    \mathbf{e}^i_v &= \mathcal{E}_v(\mathbf{I}_M^i) + \mathbf{e}_\beta \quad \text{for } i = 1,2,\ldots,N,
    \label{eq:embedding}
\end{aligned}
\end{equation}
where $\mathbf{e}_\alpha$ and $\mathbf{e}_\beta$ are learnable tokens introduced to differentiate embeddings extracted from different modalities.

\noindent\textbf{Audio-Visual Correspondence Learning.}  
To extract audio-related textual details while maintaining the identity from the reference image, we employ an Audio-Visual Alignment Unit (AVAU) to capture the interplay between audio and visual embeddings ($\{\mathbf{e}_v^i\}_{i=1}^{N}$ and $\mathbf{e}_a$). The AVAU ensures that IPTalker remains faithful to the audio input while incorporating valuable visual cues from reference images, enhancing the realism of the generated lip shapes. Specifically, the AVAU calculates the distance between each reference mouth image and the audio input, effectively identifying the reference image that most closely aligns with the target mouth shape dictated by the audio. This selection process allows the model to focus on the most relevant visual information, minimizing the potential loss of critical texture details. The structure of the AVAU is illustrated in Figure~\ref{fig:AVAU_align}. We stack multiple AVAUs within the alignment module to further fuse the different modalities:
\begin{equation}
    \{\mathbf{w}_v^i\}_{i=1}^N, \mathbf{w}_a = \mathcal{F}_{u}^k \circ \mathcal{F}_{u}^{k-1} \circ \ldots \circ \mathcal{F}_{u}^1 (\{\mathbf{e}_v^i\}_{i=1}^{N}, \mathbf{e}_a),
\end{equation}
where $\mathcal{F}_{u}^i$ represents the $i$-th AVAU. Utilizing multiple AVAUs allows our framework to fully integrate audio and visual information characteristics, thereby selecting the visual information that best matches the audio.

\noindent\textbf{Visual Information Aggregation.}  
As shown in Figure~\ref{fig:AVAU_align}, we obtain a set of audio-aware visual embeddings $\{\mathbf{w}_v^i\}_{i=1}^N$ and a visual-aware audio embedding $\mathbf{w}_a$ from multiple AVAUs. We combine these embeddings and feed the combination into a cross-modalities encoder $\mathcal{E}_{cm}$ to produce the alignment feature $\mathbf{v}_{alg}$:
\begin{equation}
    \mathbf{v}_{alg} = \mathcal{E}_{cm}(\{\mathbf{w}_v^i\}_{i=1}^{N}, \mathbf{w}_a) + \mathbf{e}_a.
\end{equation}
Here, we introduce a skip connection between the audio embedding $\mathbf{e}_a$ and the output of $\mathcal{E}_{cm}$ to emphasize audio cues. This addition benefits the subsequent generation process, as confirmed by our experiments.

\subsection{Warping Module}

Temporal consistency is paramount in talking head video generation. Existing GAN-based methods \cite{prajwal2020lip, liang2022expressive} primarily generate pixels directly guided by audio cues. Consequently, consecutive frames are generated based on differing audio cues, which can lead to temporal inconsistencies. To overcome this issue, we adopt a spatial deformation method that ensures most pixels in the generated video originate from the same source image, significantly enhancing temporal coherence.

\noindent\textbf{Reference Face Mask.}  
Our objective is to dub the mouth shape of the person in the video based on the driving audio. Therefore, we need to modify the original mouth shape according to the audio. Intuitively, we first mask the original mouth region and then inpaint the masked area.

\begin{figure}[t]
\centering
\includegraphics[width=\linewidth]{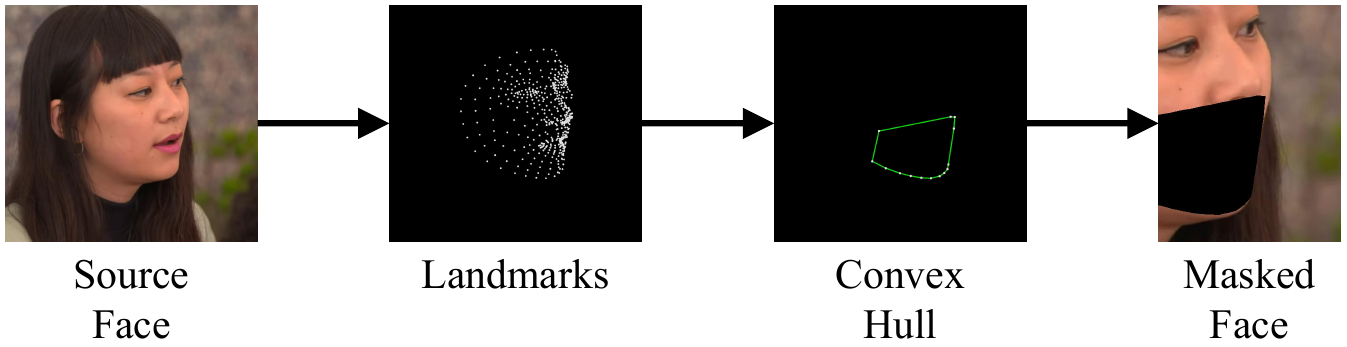}
\caption{We obtain a precise mask by calculating the convex hull of the lower-half facial landmarks set.}
\label{fig:mask}
\end{figure}

As depicted in Figure~\ref{fig:mask}, to accurately mask the mouth region of reference images, we first extract the facial landmarks of the source face using Mediapipe~\cite{lugaresi2019mediapipe}. We then compute the convex hull of the lower-half facial landmarks set, allowing us to obtain a precise mask for both frontal and profile faces.

\noindent\textbf{Feature Fusion.}  
As shown in Figure~\ref{fig:warping_inpainting}, given the source image $\mathbf{I}_S \in \mathbb{R}^{3 \times H \times W}$ with the mouth region masked, the source image encoder $\mathcal{E}_S$ extracts the source feature $\mathbf{F}_S \in \mathbb{R}^{C \times \frac{H}{4} \times \frac{W}{4}}$. Additionally, a reference image encoder $\mathcal{E}_R$ extracts the reference features $\mathbf{F}_R^i \in \mathbb{R}^{\frac{C}{N} \times \frac{H}{4} \times \frac{W}{4}}$ from $N$ reference images $\mathbf{I}_R^i \in \mathbb{R}^{3 \times H \times W}$ for $i = 1, \dots, N$:
\begin{equation}
    \begin{aligned}
        \mathbf{F}_S &= \mathcal{E}_S(\mathbf{I}_S), \\
        \mathbf{F}_R^i &= \mathcal{E}_R(\mathbf{I}_R^i) \quad \text{for } i = 1, \dots, N, \\
        \mathbf{F}_R &= \mathbf{F}_R^1 \Vert \mathbf{F}_R^2 \Vert \dots \Vert \mathbf{F}_R^N,
    \end{aligned}
\end{equation}
where $\Vert$ denotes feature concatenation. After obtaining the source feature $\mathbf{F}_S$ and the concatenated reference feature $\mathbf{F}_R$, we utilize a fusion block to aggregate these features:
\begin{equation}
    \mathbf{F}_u = \mathcal{E}_u(\mathbf{F}_S + \mathbf{F}_R),
\end{equation}
where $\mathcal{E}_u$ represents the fusion block. This aggregation effectively merges and processes features from different reference images, leading to more accurate and coherent video generation results.

\noindent\textbf{Feature Warping.}  
Our goal is to generate an accurate facial motion flow to warp the reference pixels while minimizing occlusion regions. To better preserve vivid source textures and achieve better generalization, we employ a warping network based on an encoder-decoder structure utilizing Adaptive Instance Normalization (AdaIN) \cite{huang2017arbitrary}. The warping network takes the fused features $\mathbf{F}_u$ as input and injects mouth shape and audio information from the audio-visual alignment feature embedding $\mathbf{v}_{alg}$ using the AdaIN operation before each layer.

As illustrated in Figure~\ref{fig:warping_inpainting}, the warping network first uses AdaIN to modulate the audio-visual alignment feature embedding $\mathbf{v}_{alg}$. The encoder-decoder structure processes the fused features $\mathbf{F}_u$ to produce a facial motion flow $\mathbf{M}$ conditioned on the modulated features:
\begin{equation}
    \mathbf{M} = \mathcal{F}_{\text{UNet}}(\mathbf{F}_u \, | \, \text{AdaIN}(\mathbf{v}_{alg})),
\end{equation}
where $\mathcal{F}_{\text{UNet}}$ denotes the encoder-decoder architecture used in the warping network. Subsequently, the predicted motion flow $\mathbf{M}$ is used to warp the reference features $\mathbf{F}_R$ to obtain a preliminary target feature $\mathbf{F}_w$:
\begin{equation}
    \mathbf{F}_w = \text{Warp}(\mathbf{F}_R, \mathbf{M}).
\end{equation}

\subsection{Inpainting Module}

To selectively modify only the mouth region of the source image while preserving other facial features such as the eyes and nose, the source feature $\mathbf{F}_S$ is skip-connected within the inpainting module. Specifically, the output feature from the warping module $\mathbf{F}_w$ and the source feature $\mathbf{F}_S$ are concatenated along the feature channels and fed into the SPADE \cite{park2019semantic} decoder, tasked with inpainting the masked mouth area.

To maintain semantic consistency throughout the multi-layer convolutional process within the SPADE decoder, the original features are injected using the SPADE operation before each convolutional layer. SPADE predicts normalization parameters through convolution layers based on the input features. These parameters are then used to normalize the subsequent convolution layers, ensuring that the generated content remains semantically aligned with the original facial features.

\begin{figure}[t]
\centering
\includegraphics[width=\linewidth]{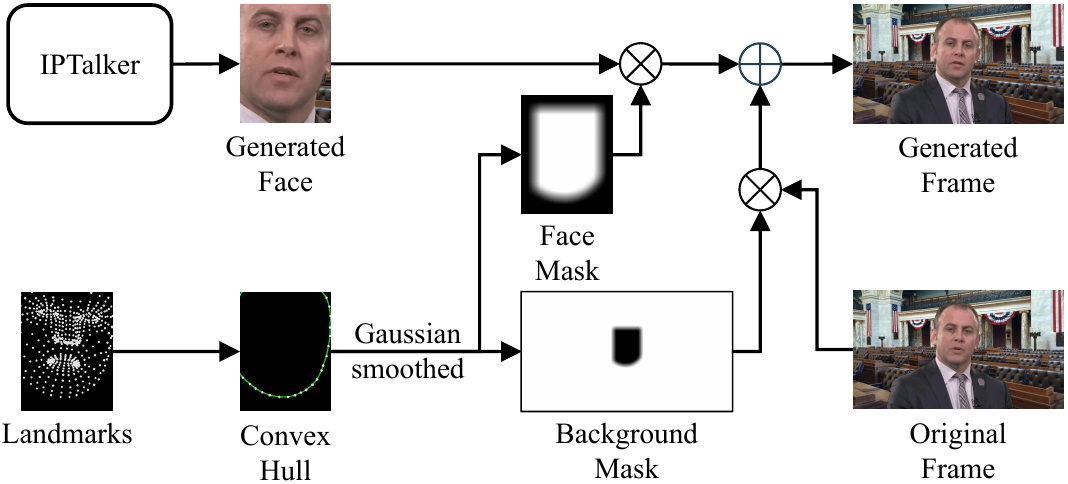}
\caption{We paste the generated face onto the original frame using a Gaussian-smoothed mask to eliminate artifacts around the facial region.}
\label{fig:paste}
\end{figure}

Occasionally, small regions with artifacts may appear around the face in generated images. To address this, we first calculate the convex hull of the entire set of facial landmarks to obtain a comprehensive mask. We then paste the generated face onto the original frame using a Gaussian-smoothed mask, as illustrated in Figure~\ref{fig:paste}.

\subsection{Loss Function}

During training, our model is optimized using a combination of three loss functions: perception loss \cite{johnson2016perceptual}, GAN loss \cite{mao2017least}, and lip-sync loss \cite{prajwal2020lip}.

\noindent\textbf{Perception Loss.}  
Following \cite{zhang2023dinet}, we compute the perception loss at two image scales. Specifically, we downsample the output image $\mathbf{I}_O \in \mathbb{R}^{3 \times H \times W}$ and the real image $\mathbf{I}_r \in \mathbb{R}^{3 \times H \times W}$ to obtain $\mathbf{\hat{I}}_O \in \mathbb{R}^{3 \times \frac{H}{2} \times \frac{W}{2}}$ and $\mathbf{\hat{I}}_r \in \mathbb{R}^{3 \times \frac{H}{2} \times \frac{W}{2}}$, respectively. Paired images $\{\mathbf{I}_O, \mathbf{I}_r\}$ and $\{\mathbf{\hat{I}}_O, \mathbf{\hat{I}}_r\}$ are input into a pre-trained VGG-19 network to compute the perception loss:
\begin{equation}
    \mathcal{L}_{p} = \sum_{i=1}^{N} \frac{\| V_i(\mathbf{I}_O) - V_i(\mathbf{I}_r) \|_1 + \| V_i(\mathbf{\hat{I}}_O) - V_i(\mathbf{\hat{I}}_r) \|_1}{2NW_iH_iC_i},
\end{equation}
where $V_i(\cdot)$ denotes the activation at layer $i$ of the VGG-19 network, and $W_i$, $H_i$, and $C_i$ represent the width, height, and number of channels at layer $i$.

\noindent\textbf{GAN Loss.}  
We employ the Least Squares GAN (LS-GAN) loss \cite{mao2017least} in our method. The GAN loss is defined as:
\begin{equation}
    \mathcal{L}_{GAN} = \mathcal{L}_{D} + \mathcal{L}_{G},
\end{equation}
where
\begin{equation}
    \mathcal{L}_{D} = \frac{1}{2} \mathbb{E}\left[ (D(\mathbf{I}_r) - 1)^2 \right] + \frac{1}{2} \mathbb{E}\left[ (D(\mathbf{I}_O) - 0)^2 \right],
\end{equation}
\begin{equation}
    \mathcal{L}_{G} = \mathbb{E}\left[ (D(\mathbf{I}_O) - 1)^2 \right],
\end{equation}
with $D(\cdot)$ representing the discriminator.

\noindent\textbf{Lip-Sync Loss.}  
To enhance the synchronization of lip movements in dubbed videos, we incorporate a lip-sync loss \cite{prajwal2020lip}. The lip-sync loss is defined as:
\begin{equation}
    \mathcal{L}_{sync} = \mathbb{E}\left[ ( \text{SyncNet}(\mathbf{A}, \mathbf{I}_O) - 1 )^2 \right],
\end{equation}
where $\text{SyncNet}(\cdot)$ is a pre-trained network that evaluates lip synchronization between the audio $\mathbf{A}$ and the output image $\mathbf{I}_O$.

\noindent\textbf{Total Loss.}  
The final loss $\mathcal{L}$ is a weighted sum of the perception loss, GAN loss, and lip-sync loss:
\begin{equation}
    \mathcal{L} = \lambda_{p} \mathcal{L}_{p} + \lambda_{sync} \mathcal{L}_{sync} + \mathcal{L}_{GAN},
\end{equation}
where $\lambda_{p}$ and $\lambda_{sync}$ are hyperparameters controlling the contributions of $\mathcal{L}_{p}$ and $\mathcal{L}_{sync}$, respectively. In our experiments, we set $\lambda_{p} = 10$ and $\lambda_{sync} = 0.1$.

\begin{figure*}[!t]
\centering
\includegraphics[width=\textwidth]{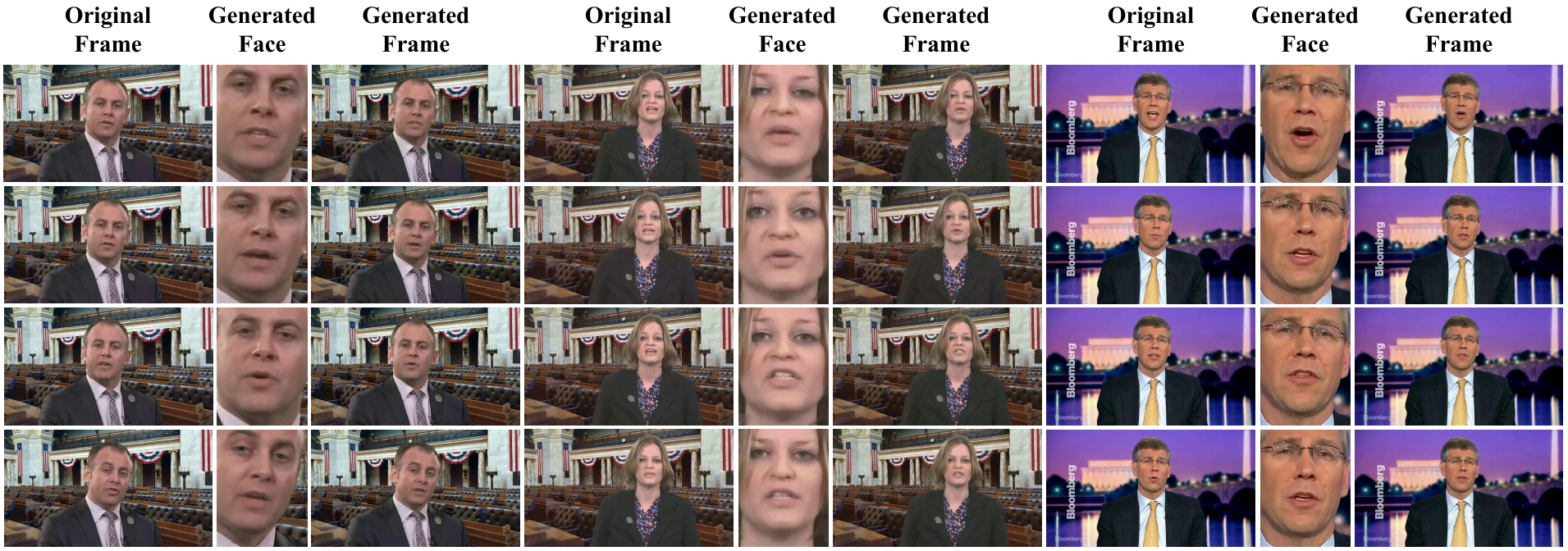}
\caption{Synthetic results of our method. IPTalker generates high-resolution images with precise mouth shapes and realistic faces while effectively preserving the reference mouth texture details.}
\label{fig:result}
\end{figure*}

\section{Experiment}
In this section, the two datasets used in the experiment, HDTF dataset \cite{zhang2021flow} and VFHQ dataset \cite{xie2022vfhq}, will be first introduced, and then the data processing process will be explained. Next, the excellent performance of our method is verified by showing the synthetic results. Then, the intermediate results of the model are visualized to prove the effectiveness of the warping module. Next, a large number of qualitative and quantitative experiments are carried out to compare the proposed method with other state-of-the-art works. At the end of this section, we conduct ablation studies.

\subsection{Datasets and Settings}
We conduct experiments on two high-resolution talking face datasets: HDTF dataset \cite{zhang2021flow} and VFHQ dataset \cite{xie2022vfhq}.

\noindent\textbf{HDTF dataset} focuses on high-quality facial videos. It contains about 300 videos in 720p or 1080p resolution, mostly lectures or TV hosting, and most of them are facing forward and have little head movement. In the experiment 30 videos were randomly selected for testing.

\noindent\textbf{VFHQ dataset} contains about 8,000 high-resolution videos from more than 20 different countries and in a variety of languages. One of the distinguishing features of the dataset is that a large number of videos show the face facing sideways, and the amplitude of head movement is relatively high in these videos. In the experiment, 20 videos were randomly selected for testing.

In data preprocessing, the audio is first extracted from the video, and then the 29 dimension audio features are extracted using the pre-trained deepspeech model. Next, resample all the videos in 25fps. Mediapipe is then used to extract 468 facial landmarks. The facial region is cropped based on the facial landmarks and resized into 416×320 resolution. Finally, for each cropped face image, we obtain the landmarks list that can cover the largest area of the lower-half face by calculating convex hull.
\subsection{Metrics}
In this work, we apply several metrics to quantitatively evaluate our methods: Structural Similarity Index Measure (SSIM), Peak Signal-to-Noise Ratio (PSNR), Learned Perceptual Image Patch Similarity (LPIPS), Lip-Sync Evaluation Confidence (LSE-C), and Lip-Sync Evaluation Distance (LSE-D). SSIM mainly quantifies the similarity between two images and it considers changes in structural information, luminance and contrast. PSNR is derived from the Mean Squared Error (MSE) between two images and it measure the similarity btween the generated video and ground truth video in pixl-level space. LPIPS is a metric designed to measure perceptual similarity between images based on deep neural network features. Specifically, LPIPS computes the distance between feature representations of two images extracted from a pretrained Vgg network~\cite{simonyan2014very}. The LSE-C and LSE-D are the evaluation metric for lip-syncing.

\subsection{Synthetic Results}
Figure~\ref{fig:result} shows the synthesized results of our method. We display the original frames, the generated faces and the generated frames. Our method can generate high-resolution images with precise mouth shapes and realistic face while preserving the reference mouth texture details well.

Visualizing the features gives us a deeper understanding of how the model works and how much detail it can handle during the generation process. As shown in Figure~\ref{fig:warping_inpainting}, the average feature map of $\mathbf{F}_w$ is displayed as a picture, and it can be found that the mouth shape is close to the generated image, which demonstrates the effectiveness of the warping module.

\begin{table*}[]
\caption{\label{tab:vssota} Quantitative comparisons with the state-of-the-art methods on talking head generation. }
\centering

{
\small
\centering
\begin{tabular}{lccccccccccc}
\hline
                    & \multicolumn{5}{c}{\textbf{HDTF}}          &  & \multicolumn{5}{c}{\textbf{VFHQ}}  \\ \cline{2-6} \cline{8-12} 
              & SSIM \(\uparrow\) & PSNR \(\uparrow\) & LPIPS\(\downarrow\) & LSE-C\(\uparrow\) &  LSE-D\(\downarrow\) &  & SSIM\(\uparrow\) &PSNR\(\uparrow\) & LPIPS\(\downarrow\) & LSE-C\(\uparrow\) & LSE-D\(\downarrow\)  \\ \hline
DreamTalk & 0.3701 & 20.2403 & 0.1661 & 6.9871 & 7.8541 & & 0.2845 & 19.7338 & 0.2205 & 5.3107 & 8.5509 \\
MakeItTalk & 0.1892 & 17.3801 & 0.2271 & 5.2611 & 9.7741 & & 0.1773 & 17.2889 & 0.2813 & 3.8217 & 10.2887  \\
Audio2Head & 0.1015 & 16.0437 & 0.2478 & 7.8821 & 7.2281 & & 0.0715 & 13.7549 & 0.4018 & 5.5181 & 8.3107 \\
SadTalker & 0.1570 & 16.7765 & 0.2436 & 8.0813 & 11.2647 & & 0.1474 & 16.6458 & 0.2601 & 5.4164 & 8.5383      \\
Hallo & 0.1070 & 17.3208 & 0.3066 & 7.8140 & 7.5877 & & 0.1014 & 16.0653 & 0.3797 & \textbf{6.5482} & \textbf{8.0465} \\
AniPortrait & 0.1089 & 17.3683 & 0.2901 & 0.2566 & 14.4621 & & 0.0782 & 16.1488 & 0.3687 & 0.9249 & 13.0160 \\
DINet & 0.5592 & 27.7341 & 0.0804 & 7.8775 & 7.3757 &  & 0.4558 & 23.3589 & 0.1761 & 4.6666 & 9.0250  \\\hline
Ground Truth & 1.0000 & N/A & 0.0000 & 8.8577 & 6.5751 &  & 1.0000 & N/A &  0.0000 & 7.0019 & 7.4634      \\
Ours & \textbf{0.6365} & \textbf{30.8796} & \textbf{0.0582} & \textbf{8.2741} & \textbf{7.0218} & & \textbf{0.5992}  & \textbf{28.6386} & \textbf{0.0705} & 6.0691 & 8.6708 \\ \hline
\end{tabular}
}
\end{table*}

\subsection{Comparisons with State-of-the-Art Works}

We compare our method with state-of-the-art open-source talking head models, including DreamTalk \cite{ma2023dreamtalk}, MakeItTalk \cite{zhou2020makelttalk}, Audio2Head \cite{wang2021audio2head}, SadTalker \cite{zhang2023sadtalker}, Hallo \cite{xu2024hallo}, AniPortrait \cite{wei2024aniportrait}, and DINet \cite{zhang2023dinet}.

\begin{figure*}[!t]
\centering
\includegraphics[width=\textwidth]{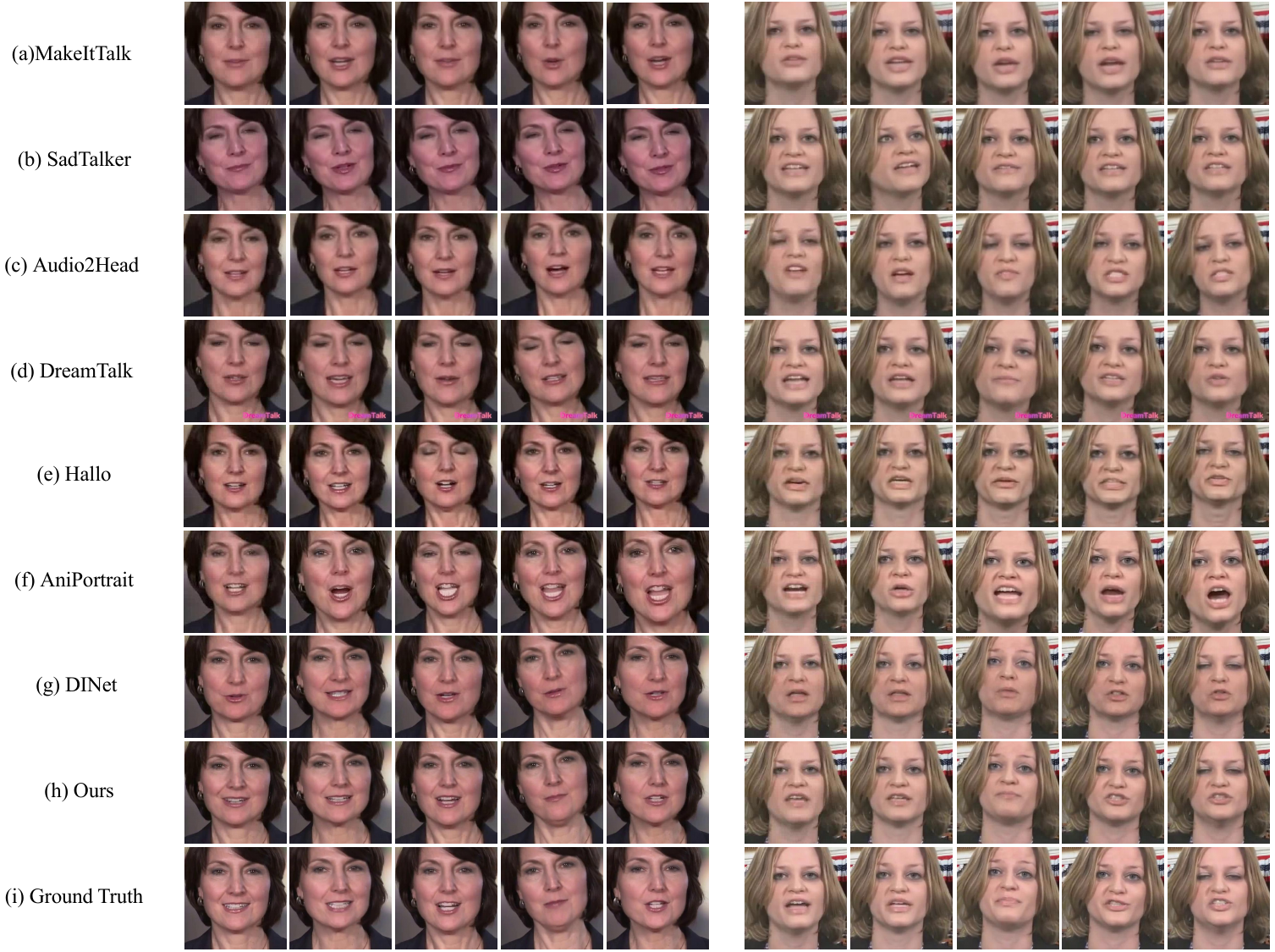}
\caption{Qualitative comparisons with state-of-the-art works. The results demonstrate that our model can generate high-resolution images with precise mouth shapes while preserving the texture details of the reference mouth effectively, with almost no artifacts.}
\label{fig:compare}
\end{figure*}

\textbf{Qualitative Comparisons.} We first perform qualitative comparisons with these state-of-the-art models. Figure~\ref{fig:compare} illustrates the results. It can be observed that existing methods such as MakeItTalk \cite{zhou2020makelttalk} and SadTalker \cite{zhang2023sadtalker} struggle to generate accurate mouth shapes that align with the audio. Sometimes, the mouth movements correspond to the audio but do not match the speaking habits of the reference identity, as shown in Figures~\ref{fig:compare}(a) and \ref{fig:compare}(b).  Other methods, like Audio2Head \cite{wang2021audio2head} and DreamTalk \cite{ma2023dreamtalk}, perform better in generating mouth shapes that somewhat match the audio. However, the image clarity and the preservation of texture details in the reference mouth are not satisfactory, as depicted in Figures~\ref{fig:compare}(c) and \ref{fig:compare}(d). Diffusion model-based methods such as Hallo \cite{xu2024hallo} and AniPortrait \cite{wei2024aniportrait} can generate high-resolution images with rich details. Nevertheless, the details do not always align with the reference image, and severe deformations and artifacts may appear in the mouth and teeth regions, as shown in Figures~\ref{fig:compare}(e) and \ref{fig:compare}(f). Our method calculates the distance between the audio and various reference images to identify the reference mouth shape that best matches the audio. This approach allows the model to focus more effectively on relevant features. Compared to other existing methods, our unique design enables the generation of high-resolution images with precise mouth shapes while preserving the texture details of the reference mouth, as demonstrated in Figure~\ref{fig:compare}(i). Additionally, our method produces almost no artifacts, as shown in Figure~\ref{fig:compare}(h).

\textbf{Quantitative Comparisons.} Table~\ref{tab:vssota} presents the results of quantitative comparisons. To evaluate visual quality, we compute the metrics of Structural Similarity (SSIM) \cite{wang2004image}, Peak Signal-to-Noise Ratio (PSNR), and Learned Perceptual Image Patch Similarity (LPIPS) \cite{zhang2018unreasonable}. For audio-visual synchronization, inspired by \cite{prajwal2020lip}, we calculate Lip Sync Error Confidence (LSE-C) and Lip Sync Error Distance (LSE-D).

In terms of visual quality, our method achieves the best results across all metrics. Regarding audio-visual synchronization, our method performs best on the HDTF dataset. The slight underperformance on the VFHQ dataset can be attributed to training exclusively on the HDTF dataset, which primarily consists of front-facing videos, whereas VFHQ includes a significant number of profile-face videos. 

\noindent\textbf{Identity Fine-tuning.} We have demonstrated the performance of our model on specific identity fine-tuning through experiments on the HDTF dataset. Specifically, each video in the test set is divided into two segments on average. For each reference identity, the first part of the video is used for inference, and the second part is used for fine-tuning. Table~\ref{tab:finetune} shows quantitative results of fine-tuning with 0, 100, 200, and 300 steps on the HDTF dataset, where 0 steps indicate no fine-tuning. The results indicate that identity fine-tuning significantly enhances the generative capability of our model. As the number of training steps increases, the visual metrics improve. And the optimal number of training steps to enhance lip synchronization metrics and prevent overfitting is about 200 steps.

\begin{table}[htb]
\caption{\label{tab:finetune} Quantitative Results of Identity Fine-tuning on the HDTF Dataset.}
\centering

\begin{tabular}{lccccc}
\hline
Steps & SSIM \(\uparrow\) & PSNR \(\uparrow\) & LPIPS \(\downarrow\) & LSE-C \(\uparrow\) & LSE-D \(\downarrow\) \\ \hline
0    & 0.6332 & 30.8919 & 0.0584  & 8.2349 & 7.0207 \\
100  & 0.6688 & 33.7399 & 0.0388  & 8.2464 & 7.0027 \\
200  & 0.6724 & 34.0838 & 0.0370  & \textbf{8.2652} & \textbf{6.9742} \\
300  & \textbf{0.6747} & \textbf{34.2662} & \textbf{0.0364} & 8.2624 & 6.9888 \\ \hline
\end{tabular}
\end{table}

\subsection{Ablation Study}

In this section, we conduct ablation studies on the HDTF dataset to validate the impact of each component in our method. Specifically, we examine four conditions:

\begin{enumerate}
    \item \emph{Ours w/o Alignment}: We remove the alignment module and use a multi-layer convolutional encoder to extract audio features, which are subsequently injected into the warping network.
    \item \emph{Ours w/o SPADE}: We remove the SPADE decoder and instead utilize a multi-layer convolutional decoder to inpaint the image.
    \item \emph{Ours w/o $\mathcal{E}_{cm}$}: We eliminate $\mathcal{E}_{cm}$ after the transformer encoder and directly inject the output token into the warping network.
    \item \emph{Ours}: Our proposed method.
\end{enumerate}

\begin{figure}[!t]
\centering
\includegraphics[width=0.9\linewidth]{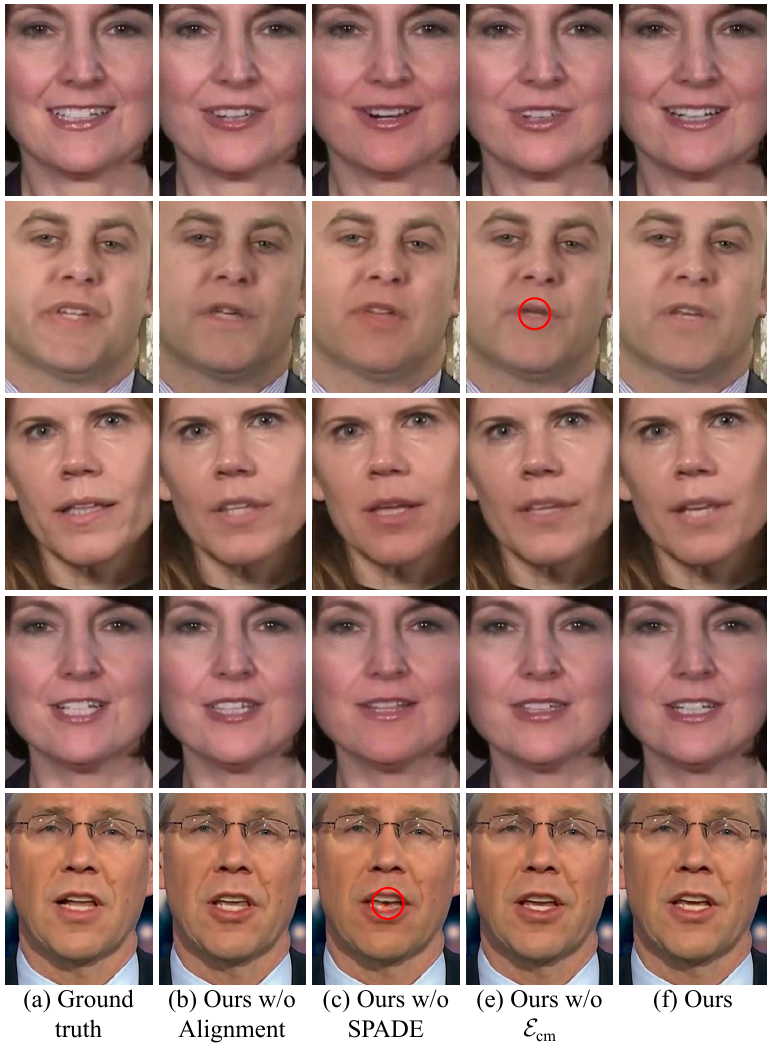}
\caption{Qualitative results of the ablation study. We conduct ablation studies on the HDTF dataset to validate the effects of the alignment module, SPADE decoder, and $\mathcal{E}_{cm}$ in our model.}
\label{fig:ablation}
\end{figure}

Figure~\ref{fig:ablation} illustrates the qualitative results of the ablation experiments, while Table~\ref{tab:ablation} presents the quantitative results on the HDTF dataset.

\noindent\textbf{Effectiveness of the Alignment Module.} Our model leverages the alignment module to obtain attention information. As shown in Figure~\ref{fig:ablation}(b), although the lip shape of the synthetic face image aligns with the driving audio, the mouth's texture details appear blurry. From Figures~\ref{fig:ablation}(b) and (f) and Table~\ref{tab:ablation}, we observe that our method with the alignment module excels at preserving the reference mouth texture more effectively and achieves superior results in both visual and audio-visual synchronization metrics. This improvement primarily stems from avoiding the use of a single encoder for audio features. While a single encoder can capture the correspondence between audio and mouth shapes, the model fails to preserve mouth details that match the reference identity without the alignment process. Additionally, the deformation process lacks reference attention information, causing details and textures from different reference images to intermingle, ultimately resulting in a blurred final mouth image.

\noindent\textbf{Effectiveness of the SPADE Decoder.} To preserve the warping features during the generation process, we employ a SPADE decoder in our method. As depicted in Figure~\ref{fig:ablation}(c), the deviation from the source image is most evident in this condition, especially concerning the mouth texture details, where artifacts occasionally appear. Table~\ref{tab:ablation} also shows that the SPADE decoder improves the model's visual metrics to a certain extent. These results suggest that, during the inpainting process, maintaining feature semantics is challenging without the SPADE operation, leading to significant loss of oral texture details in the generated image. Furthermore, during training, the warping network struggles to learn how to produce an effective flow field without the SPADE decoder.

\noindent\textbf{Effectiveness of the Cross-Modalities Encoder ($\mathcal{E}_{cm}$).} We incorporate a cross-modalities encoder, $\mathcal{E}_{cm}$, to aggregate all embeddings. As shown in Figures~\ref{fig:ablation}(d) and (f), our full method produces more accurate lip shapes compared to the "Ours w/o $\mathcal{E}_{cm}$" condition. The results in Table~\ref{tab:ablation} demonstrate that the model's performance improves when the cross-modalities encoder is utilized. Compared to the model without the entire alignment module, the model with AVAUs shows some improvement in the lip synchronization metric. However, the complete model with the cross-modalities encoder, which processes the output representations of AVAUs, exhibits a much more significant enhancement in lip synchronization performance. These results strongly validate the effectiveness of this module. A possible explanation is that, although the output of AVAUs contains all the necessary information, the complexity of the reference attention information is significantly less than that of the target lip shape information. By concatenating the AVAUs output and directly injecting it into the warping network, the model struggles to capture crucial lip shape information. To address this, we employ $\mathcal{E}_{cm}$ to compress the attention information and then concatenate it with the audio features as the output of the alignment module. This facilitates easier access to lip shape information for the warping module, thereby enhancing the precision of lip shapes.

\begin{table}[t]
\caption{\label{tab:ablation} Quantitative Results of Ablation Study on the HDTF Dataset.}
\centering
\resizebox{\linewidth}{!}{
    \begin{tabular}{lccccc}
    \hline
                  & SSIM \(\uparrow\) & PSNR \(\uparrow\) & LPIPS \(\downarrow\) & LSE-C \(\uparrow\) & LSE-D \(\downarrow\) \\ \hline
    Ours w/o Alignment & 0.6246 & 29.6988 & 0.0620 & 7.9878 & 7.2844 \\
    Ours w/o SPADE     & 0.6274 & 29.8337 & 0.0626 & 8.0865 & 7.2544 \\
    Ours w/o $\mathcal{E}_{cm}$ & 0.6265 & 29.7840 & 0.0627 & 8.0697 & 7.2375 \\ \hline
    Ours               & \textbf{0.6365} & \textbf{30.8796} & \textbf{0.0582} & \textbf{8.2741} & \textbf{7.0218} \\ \hline
    \end{tabular}
}
\end{table}

\section*{Limitations}

While our method excels at generating realistic facial images that preserve complex mouth texture details, it still has certain limitations. Specifically, we focus solely on spatial deformation in the mouth area, resulting in a lack of dedicated design to ensure harmonious integration with other facial features. Consequently, the generated mouth may exhibit inconsistencies in angle and position relative to the surrounding facial regions, potentially leading to a phenomenon where the mouth appears to be "floating."

\section*{Conclusion}
In this paper, we propose a talking head generation model capable of generating high-fidelity images while preserving texture details of reference mouth. Our model comprises three fundamental components: the alignment module, the warping module and the inpainting module. 
The alignment module employs attention mechanism to identify the optimal reference mouth image that closely resembles the target mouth shape corresponding to the audio input. 
The warping module is responsible for spatially deforming the reference image based on the information provided by the alignment module. 
The inpainting module then conpensate the occlusion regions which are not considered
by the motion flow and generated the final result. 
Extensive qualitative and quantitative experiments have validated the proposed method can generate high-resolution images with precise mouth shapes while preserving the reference mouth texture details better than other existing methods.

\bibliographystyle{IEEEtran}
\bibliography{ref}

\vfill

\end{document}